\documentclass[runningheads]{llncs}
\usepackage[T1]{fontenc}

\usepackage{cite}
\usepackage{graphicx}
\usepackage{wrapfig}
\usepackage{amsmath}
\usepackage{tikz}
\usepackage{makecell}
\usepackage[dvipsnames]{xcolor}
\usepackage{todonotes}
\usepackage{ifthen}
\usepackage{float}

\usepackage[subpreambles=true]{standalone}
\usepackage{import}
\providecommand{\fillStdDev}[4]{
  \pgfplotstableread[col sep=comma]{#1}\datatable
  \pgfplotstablecreatecol[
    create col/expr={\thisrow{mean_accuracy}*#3 + \thisrow{std_accuracy}*#3}
  ]{mean_plus_std}{\datatable}
  \pgfplotstablecreatecol[
    create col/expr={\thisrow{mean_accuracy}*#3 - \thisrow{std_accuracy}*#3}
  ]{mean_minus_std}{\datatable}

  \addplot[name path=upper, draw=#4!50, forget plot]
    table[x expr=(\thisrow{epoch}+1)*#2, y=mean_plus_std]{\datatable};
  \addplot[name path=lower, draw=#4!50, forget plot]
    table[x expr=(\thisrow{epoch}+1)*#2, y=mean_minus_std]{\datatable};
  \addplot[
    fill=#4, fill opacity=0.2, forget plot] fill between[
    of=upper and lower
  ];
}

\providecommand{\plotMean}[7]{%
  \pgfplotstableread[col sep=comma]{#1}\datatable

  \addplot[
    color=#4,
    draw,
    legend image post style={line width=1.5pt},
    #7
  ] table[x expr=(\thisrow{epoch}+1)*#2, y expr=\thisrow{mean_accuracy}*#3]{\datatable};

  \ifthenelse{\equal{#5}{true}}{
    \addlegendentry{#6}
  }{
  }%
}

\providecommand{\plotMeanAndStdDev}[7]{
  
  \pgfplotstableread[col sep=comma]{#1}\datatable

  \pgfplotstablecreatecol[
    create col/expr={\thisrow{mean_accuracy}*#3 + \thisrow{std_accuracy}*#3}
  ]{mean_plus_std}{\datatable}
  \pgfplotstablecreatecol[
    create col/expr={\thisrow{mean_accuracy}*#3 - \thisrow{std_accuracy}*#3}
  ]{mean_minus_std}{\datatable}

  \addplot[name path=upper, draw=#4!50, forget plot]
    table[x expr=(\thisrow{epoch}+1)*#2, y=mean_plus_std]{\datatable};
  \addplot[name path=lower, draw=#4!50, forget plot]
    table[x expr=(\thisrow{epoch}+1)*#2, y=mean_minus_std]{\datatable};
  \addplot[
    fill=#4, fill opacity=0.2, forget plot] fill between[
    of=upper and lower
  ];

  \addplot[
    color=#4,
    draw,
    legend image post style={line width=1.5pt},
    #7
  ] table[x expr=(\thisrow{epoch}+1)*#2, y expr=\thisrow{mean_accuracy}*#3]{\datatable};

  \ifthenelse{\equal{#5}{true}}{
    \addlegendentry{#6}
  }{
  }%
}

\usepackage{pgfplots}
\usepackage{pgfplotstable}
\usetikzlibrary{intersections}
\usepgfplotslibrary{fillbetween}
\pgfplotsset{compat=1.18}
\usepackage{subcaption}
\usepackage{mwe}
\usepackage{threeparttable}
\usepackage{multirow}
\usepackage{booktabs}
\usepackage{tabularx}
\usepackage{multicol}
\usepackage{multirow}
\usepackage{amsmath}

\DeclareMathOperator*{\argmin}{argmin}

\usepackage{placeins}  %
\usepackage{hyperref}
\begin{document}
\title{Predictive Coding-based Deep Neural Network Fine-tuning for Computationally Efficient Domain Adaptation}
\titlerunning{Predictive Coding-based Neural Network Fine-tuning for Domain Adaptation}
\author{Matteo Cardoni\orcidID{0000-0002-2759-5310} \and\\ Sam Leroux\orcidID{0000-0003-3792-5026}}
\institute{IDLab, Department of Information and Technology, Ghent University—imec,\\ 9052 Ghent, Belgium\\
\email{matteo.cardoni@ugent.be} (Corresponding author),\\ \email{sam.leroux@ugent.be}}
\maketitle              %
\begin{abstract}
As deep neural networks are increasingly deployed in dynamic, real-world environments, relying on a single static model is often insufficient. Changes in input data distributions caused by sensor drift or lighting variations necessitate continual model adaptation. In this paper, we propose a hybrid training methodology that enables efficient on-device domain adaptation by combining the strengths of Backpropagation and Predictive Coding. The method begins with a deep neural network trained offline using Backpropagation to achieve high initial performance. Subsequently, Predictive Coding is employed for online adaptation, allowing the model to recover accuracy lost due to shifts in the input data distribution. This approach leverages the robustness of Backpropagation for initial representation learning and the computational efficiency of Predictive Coding for continual learning, making it particularly well-suited for resource-constrained edge devices or future neuromorphic accelerators. Experimental results on the MNIST and CIFAR-10 datasets demonstrate that this hybrid strategy enables effective adaptation with a reduced computational overhead, offering a promising solution for maintaining model performance in dynamic environments.

\end{abstract}

\section{Introduction}
The Backpropagation (BP) algorithm \cite{rumelhart1986learning} is currently the predominant approach for training deep neural network architectures. It is, however, not without its limitations. BP requires the computation and transmission of global error signals which can be computationally intensive, apart from being biologically implausible. This has sparked interest in alternative training techniques such as those based on Predictive Coding (PC) \cite{rao_and_ballard1999PredictiveCoding, friston2005atheory, friston2009predictive_coding_free_energy}. While most of the works in this area are grounded in theoretical neuroscience, PC is also promising from an engineering point of view as it provides a much more efficient hardware-friendly training procedure for neuromorphic systems. Unlike BP, PC relies on local computations and error-driven updates that align well with the distributed, event-driven nature of neuromorphic architectures. This makes it a compelling candidate for developing scalable, low-power learning algorithms suitable for real-time applications on energy-constrained edge devices \cite{millidge2022predictivecodingfuturedeep}.

\begin{figure}
    \centering
    \begin{adjustbox}{width=\textwidth}
        \import{}{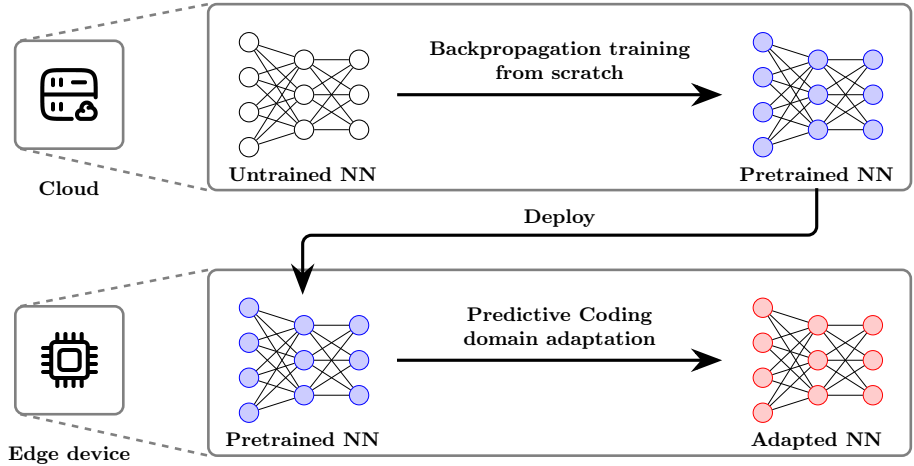}
    \end{adjustbox}
    \caption{Schematic overview of our proposed approach. After training a model offline using Backpropagation (BP), we employ Predictive Coding (PC) for on-device domain adaptation.}
    \label{fig:schematic_overview}
\end{figure}

Despite the progress made in recent years, models trained using PC still fall short of the accuracy levels achieved by those trained with BP, particularly when applied to deeper architectures and more complex datasets \cite{goemaere2025erroroptimizationovercomingexponential, pinchetti2025benchmarkingpredictivecodingnetworks}. In this paper, we propose a novel hybrid approach that combines the strengths of both algorithms. As illustrated in Figure \ref{fig:schematic_overview}, we introduce a two-stage training and deployment pipeline. In the first stage, a model is trained offline using BP on high-performance cloud infrastructure with virtually unlimited computational resources. This allows us to reach an accuracy level that would be currently impossible  to achieve using PC alone.\\
The trained model is then deployed on a resource-constrained edge device such as a wearable or sensor node. Over time, shifts in the input data distribution due to sensor drift, environmental changes, or user behavior may degrade model performance, necessitating continual adaptation. To address this, we propose using PC as a lightweight, local learning mechanism that allows efficient on-device updates, without the need for expensive BP or communication with the cloud.

The remainder of this paper is structured as follows: we provide an overview of the related work in Section \ref{sec:related_works} and describe our experimental setup and results in Section \ref{sec:experiments_and_results}. We conclude in  Section \ref{sec:conclusion} and give some pointers for future research directions in this area. 

\section{Related work}
\label{sec:related_works}
This section provides a basic overview of Backpropagation and Predictive Coding for neural network training. We refer to \cite{vanzwol2024predictivecodingnetworksinference} for a more in-depth tutorial.

\subsection{Neural networks}
A feedforward neural network (FNN) consists of a stack of $L$ layers, indexed as $\{l_0, l_1, \dots, l_L\}$, with each layer defined by its weight matrix $\theta_l$. Each layer $l$ computes an activity $a_l$ transforming its input activity $a_{l-1}$ as follows:
\begin{equation}
    a_l = f(a_{l-1}, \theta_{l})
\label{eq_forward_pass_layer_operation}
\end{equation}

This transformation depends on the type of layer (e.g. fully connected, convolutional, ...) and typically includes a non-linear activity. Setting the activation $a_0$ to a datapoint $x$, the neural network performs a function NN: $X \rightarrow Y$:

\begin{equation}
    \text{NN}(x) = ( l_L \circ l_{L-1} \circ l_{L-2} \circ ... \circ l_1)(x)
\label{eq_forward_pass_nn}
\end{equation}

The weights of the model $\theta$ are initialized randomly and tuned using a training procedure to achieve the desired behavior.

\subsection{Backpropagation}
Backpropagation (BP) is the most widely used technique for training neural networks. It aims to minimize a global loss function $\mathcal{L}$ that quantifies the discrepancy between the network's output and the desired target output. Common loss functions include Mean Squared Error (MSE) for regression tasks and Cross-entropy loss for classification tasks. The BP algorithm is able to calculate a suitable set of weights $\hat{\theta}$ numerically using gradient descent:
\begin{equation}
    \hat{\theta} = \argmin_{\theta} \mathcal{L(\theta)}
\end{equation}
In BP, the error of each layer is the gradient of the loss with respect to the activation of that layer. Because this depends on the gradients of subsequent layers, this backward computation must proceed sequentially from the output layer to the input layer. As a result, the gradient computation across layers is inherently sequential.

\subsection{Predictive Coding}
\label{subsec:predictive_coding}
Predictive Coding (PC) training for deep neural network architectures differs fundamentally from Backpropagation (BP) in that it only uses local information to perform weight updates. During training, an input sample $x$ is clamped to the first layer activity ($a_0$) and the last layer ($a_L$) is clamped to the desired output $y$. We then repeat a two-step training procedure:
\begin{enumerate}
    \item \textbf{Inference step:} In this step, the weights \( \theta_l \) are held fixed. Each layer predicts the activity of the next layer:
    \begin{equation}
        \mu_l = f(a_{l-1}, \theta_{l})
    \end{equation}
    The predictions $\mu_L$ are then used to calculate a layer-wise prediction error that measures the discrepancy between the actual and predicted activity:
    \begin{equation}
        \varepsilon_l = a_l - \mu_l
    \end{equation}
    
    Unlike during BP-based training, the actual activity of the layers is not obtained by passing the input through the model, as in Equation \ref{eq_forward_pass_layer_operation}. It is, instead, obtained by minimizing the energy function, which is the sum of the layer-wise prediction errors:
    \begin{equation}
        E(a,\theta) = \displaystyle\frac{1}{2}\sum_l(\varepsilon_l)^2
        \label{eq_global_energy_function}
    \end{equation}
    The updated values of the activities are obtained as the minimum of the energy function:
    \begin{equation}
        \hat{a}_l = \argmin_{a_l} E(a, \theta)
    \end{equation}

    The $\hat{a}_l$ are typically obtained via local gradient descent of Equation \ref{eq_global_energy_function}, calculating the update of $a_l$ as:
    \begin{equation}
        \Delta a_l = -\gamma \displaystyle\frac{\partial E}{\partial a_l} = -\gamma \Big(\varepsilon_l - (\theta_l)^T \varepsilon_{l+1} \odot f'(\theta_la_l)\Big),
        \label{eq_delta_a}
    \end{equation}
    with $\gamma$ being the \textit{inference rate}: a step size required by the gradient descent.
    This step infers the suitable activity values for the different layers, given the clamped input and target values and is typically performed by running gradient descent for a fixed number of iterations.

    It is to be noted that the dependency of $\Delta a_l$ from $\varepsilon_l$ and $\varepsilon_{l+1}$ allows $\varepsilon_L$, the error between the classification layer activity $a_L$ and the label $y$, to be propagated from the output layer $L$ to the input. Moreover, this computation can be layer-wise parallelized. 

    \item \textbf{Weight update step:} The activities, after being updated, are now held fixed, and the weights are updated to further reduce the energy function from Equation \ref{eq_global_energy_function}. The updated weights values are obtained as the minimum of the energy function: 
    \begin{equation}
        \hat{\theta}_l = \argmin_{\theta_l} E(\hat{a}, \theta)
        \label{eq:pc_weights_update}
    \end{equation}
    The updates of $\theta_l$ are obtained via local gradient descent:
    \begin{equation}
        \Delta \theta_l = -\alpha \displaystyle\frac{\partial E}{\partial \theta_l} = \alpha\varepsilon_{l+1} \odot f'(\theta_l\hat{a}_l)(\hat{a}_l)^T,
        \label{eq_delta_theta}
    \end{equation}
    with $\alpha$ being the weights learning rate. This computation too can be layer-wise parallelized.
\end{enumerate}

\subsection{Backpropagation-Predictive Coding equivalence at test time}
At test time, the activity of each layer in the Predictive Coding Network model, including the last layer, can be obtained by clamping the lowest layer to the input data and by computing updates of the activity by repeating the inference procedure. However, with the output unclamped, this can also be computed in a single pass by simply performing the forward pass of a normal neural network \cite{vanzwol2024predictivecodingnetworksinference}. This observation that models trained using Backpropagation and Predictive Coding are equivalent at test time is what motivated this work as it allows switching between both training algorithms while maintaining a compatible model.

\subsection{Applications of Predictive Coding}
Predictive Coding has seen growing application in recent years across a range of increasingly complex tasks. Initial studies focused on classification, associative memory, and denoising using Multi-Layer Perceptron (MLP) architectures \cite{whittington2017an_approximation_of_the_error_backpropagation_algorithm_in_a_pcn, pinchetti2025benchmarkingpredictivecodingnetworks, Sun2020APN}. Building on these foundations, more recent work has extended PC to convolutional neural networks (CNNs) for higher-level visual tasks. For example, \cite{wen2018deeppredictivecodingnetwork, han2018deeppredictivecodingnetwork} introduced recurrent variants of PC networks that achieved classification performance comparable to Backpropagation on the CIFAR-10 dataset, and promising, though less competitive, results on ImageNet. Furthermore, \cite{pinchetti2025benchmarkingpredictivecodingnetworks} systematically benchmarked PC training in both MLPs and CNNs for classification and associative memory tasks. This work also introduced a JAX-based implementation framework \cite{jax2018github}, which forms the basis for the PC models used in our work.

\subsection{Domain adaptation and on-device learning}
Classical artificial intelligence approaches typically assume that the training and test data are drawn from the same underlying distribution. However, this assumption often breaks down in real-world applications, particularly when deploying pretrained neural networks on devices that process data from dynamic, non-stationary environments \cite{farahani2021a_brief_review_of_domain_adaptation, SUN2015A_survey_of_multi-source_domain_adaptation}. Domain adaptation seeks to address this mismatch by adapting models trained on a source domain to perform well on a target domain with a different marginal data distribution.

Domain adaptation is particularly important in scenarios where a model is deployed on a physical device operating in a specific environment. These edge devices typically have limited computational and energy resources, which preclude the use of large, resource-intensive models. Instead, a lightweight, location-specific model, adapted to the unique characteristics of the local sensor or environment, can achieve performance comparable to that of a larger, general-purpose model, while being significantly more energy-efficient and responsive \cite{leroux2022automated}. While it is possible to perform the domain adaptation remotely, on cloud infrastructure, a full on-device approach is favorable in terms of privacy, communication overhead and system cost \cite{Zhao2022a_survey_of_dl_on_mobile_devices}. Novel algorithms that can perform this training more efficiently are therefore of high value for advancing adaptive intelligence at the edge. To the best of our knowledge, this work is the first to consider Predictive Coding for on-device domain adaptation.

\section{Experiments and results}
\label{sec:experiments_and_results}

\subsection{Experimental setup}
\label{subsec:experimental_setup}
We report results on the well-known MNIST \cite{mnist} and CIFAR-10 \cite{cifar10} image classification benchmarks. Following the setup in \cite{pinchetti2025benchmarkingpredictivecodingnetworks}, we use a three-layer fully connected neural network for MNIST, and a series of convolutional architectures inspired by the VGG family \cite{simonyan2015deepconvolutionalnetworkslargescale} for CIFAR-10. Full model specifications and training hyperparameters are provided in the Appendix.\\
All experiments were implemented using JAX \cite{jax2018github}. Models trained with Backpropagation (BP), along with their training procedures, were implemented in Equinox \cite{kidger2021equinox}. Instead, models trained with Predictive Coding (PC), along with their training procedures, were implemented in the PCX library \cite{pinchetti2025benchmarkingpredictivecodingnetworks}, which generalizes the gradients computation from Equations \ref{eq_delta_a} and \ref{eq_delta_theta} by means of the JAX automatic differentiation.\\
Each model was initially trained from scratch using BP on the original dataset (the details of these trainings are reported in the Appendix).
To simulate domain shifts, the training and test images were then transformed in three different ways, for three different scenarios:
\begin{itemize}
    \item Inversion (180 degrees rotation).
    \item Clockwise rotation of 20 degrees.
    \item Addition of random uniform noise in the range of [-0.05, 0.05] per channel (while the pixel values are in the range [0, 1]).
\end{itemize}
We evaluated the performance of both BP and PC in adapting the pretrained models to this transformed data, assessing their ability to recover lost accuracy under this input perturbation. The domain adaptation took place retraining on the entire training dataset after applying the corresponding transformation. Figure \ref{fig:retraining_procedure} visualizes the procedure, for the case where a VGG5 model (see Appendix for details) is used for training and domain adaptation on the CIFAR-10 dataset. All experiments were conducted on a cloud infrastructure.\\

\begin{figure}
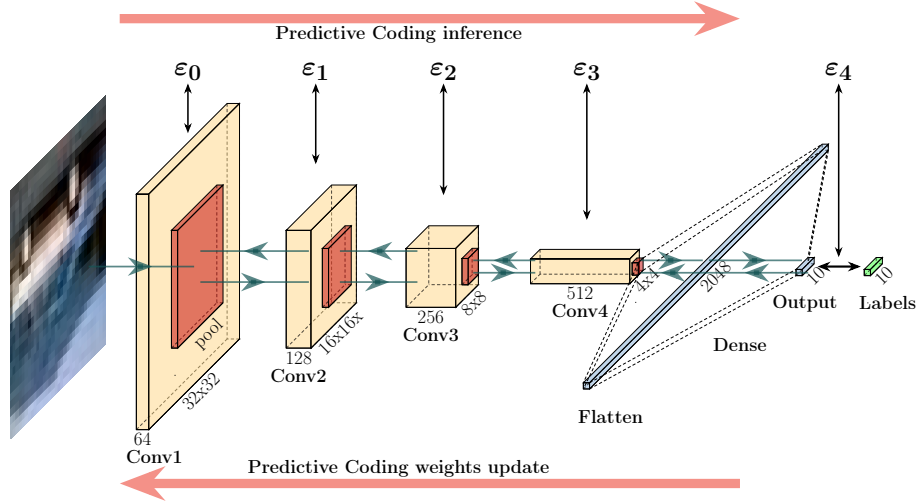

  \centering
   \begin{adjustbox}{width=\textwidth}
        \import{}{images_backprop_training.tex}
    \end{adjustbox}
  
  \begin{adjustbox}{width=\textwidth}
        \import{}{images_pred_cod_training.tex}
  \end{adjustbox}
  \caption{Experiment workflow, including the initial training of a VGG5 (see Appendix for the architecture detail) for the CIFAR-10 dataset in its original version. The same Neural Network is subsequently retrained with a domain-shifted version of the dataset. In this case, an image inversion (180 degrees rotation) has been applied to all the images.
  Opposite to the Backpropagation (BP), Predictive Coding (PC) training is based on layer-wise errors $\varepsilon_l$, that allows to compute local gradient of the weights, as in Equation \ref{eq_delta_theta}}
  \label{fig:retraining_procedure}
\end{figure}

\subsection{Results}

Table \ref{tab:train_time_mean_std} reports the average training time per epoch for both Backpropagation (BP) and Predictive Coding (PC), measured on a GeForce GTX 1080 Ti GPU. To obtain an accurate comparison, only the training time, after data loading, has been measured. Since the data transformations simulate domain shifts in data sampled by a sensor, also the data transformation has been excluded from the measurement time.\\
The results show that for the convolutional models, PC achieves a substantial speed advantage, with an average training time that is only 52\% of that required by BP. In contrast, for the MLP model, the PC training times are on average 56\% higher, likely due to the simplicity of the architecture, which limits the potential gains in efficiency from PC.

\begin{table}[h!]
 \newcommand{\accWithVar}[2]{
        #1 $\pm$ #2
     }
    \centering
    \caption{Wall-clock time per epoch (in seconds) for both Backpropagation (BP) and Predictive Coding (PC), averaged over 10 epochs for the MLP model and over 100 epochs for the VGG models. Values are reported as mean $\pm$ standard deviation. For the CNN-based architectures, PC is on average twice as fast.}
    \label{tab:train_time_mean_std}
    \begin{tabular}{p{2cm}p{2cm}p{2cm}p{2cm}}
        \toprule
        \bf Dataset & \bf Model & {\bf BP} & {\bf PC} \\
        \midrule
        MNIST & MLP & \accWithVar{1.88}{0.29} & \accWithVar{2.93}{3.17} \\
        \midrule
        \multirow{3}{*}{CIFAR10}
            & VGG5 & \accWithVar{9.48}{0.28} & \accWithVar{4.34}{1.48} \\
            & VGG7 & \accWithVar{10.48}{0.88} & \accWithVar{5.16}{2.06} \\
            & VGG9 & \accWithVar{8.48}{0.58} & \accWithVar{5.43}{1.79} \\        
        \bottomrule
    \end{tabular}
\end{table}

Although these results demonstrate that PC can reach a higher training throughput compared to BP, this does not necessarily mean that it results in faster training (i.e. performance increase over time), since accuracy gains per epoch vary across models and under different domain shifts. To investigate this, Figure \ref{fig:accuracy_vs_time} shows the models test accuracy on the domain shifted input data as a function of the training time. We compare our PC-based domain adaptation approach with a BP-based approach. As a baseline, we also include the performance for a PC and a BP model, trained from scratch on the domain-shifted data. All training details can be found in the Appendix.\\
The graphs show both the mean accuracies (middle lines) and standard deviations (color-filled areas) resulting from 5 different executions. In some of the plots the standard deviation is hardly noticeable, due to the small variance over different runs.\\
For the inverted data, the BP training from scratch baseline consistently obtains the highest accuracy. For the VGG5 model, our proposed PC-based domain adaptation achieves a similar accuracy (87.7\% $\pm$ 0.2\% compared to 88.2\% $\pm$ 0.2\%).
The VGG5 model almost reaches the accuracy obtained by the BP from scratch training, but with a much faster training, outperforming the BP-based adaptation (87.2\% $\pm$ 0.3\%) and PC-based training from scratch (86.5\% $\pm$ 0.1\%). These values correspond to the peak accuracy observed within 350 seconds of training.
The two other domain shifts proved to be more challenging, resulting in lower accuracies overall. Here, all models benefit from pretraining combined with domain adaptation.  For the VGG5 model, PC-based domain adaptation matches or surpasses the accuracy obtained with BP domain adaptation for rotated and noisy data, while it obtains a slightly lower accuracy on the VGG7 model.\\
The models trained from scratch using  PC obtain significantly lower accuracies compared to their counterparts trained using BP.
\begin{figure}[!b]
    \centering
    \begin{subfigure}{\textwidth}
        \hspace{-5mm}
        \begin{subfigure}[t]{0.38\textwidth}
            \centering
                \begin{tikzpicture}
    \begin{axis}[
      title={Inversion},
      xlabel style={at={(axis description cs:0.5, -0.1)},anchor=north},
      xlabel={Training time [s]},
      ylabel={Test accuracy (\%)},
        every axis y label/.style={
            at={(ticklabel cs:0.5, 1)},rotate=90,anchor=center,
        },
      grid=major,
      xtick={0, 100, 200, 300},
      width=\textwidth, %
      height=5cm,
      xmax=400,
      ymax=92,
      ymin=48,
      legend pos=south east,
      legend style={legend columns=2,
            /tikz/every even column/.append style={column sep=0.5cm}},
        ]
      \fillStdDev{results_flipped_vgg5_backprop_acc_results_accuracy_mean_std.csv}{9.4796}{100}{red}
      \fillStdDev{results_flipped_vgg5_backprop_fine_tune_results_accuracy_mean_std.csv}{9.4796}{100}{ForestGreen}
      \fillStdDev{results_flipped_vgg5_pred_code_after_backprop_results_accuracy_mean_std.csv}{4.3399}{100}{blue}
      \fillStdDev{results_flipped_vgg5_pred_code_standalone_results_accuracy_mean_std.csv}{4.3399}{100}{orange}
      
      \plotMean{results_flipped_vgg5_backprop_acc_results_accuracy_mean_std.csv}{9.4796}{100}{red}{false}{BP from scratch}{thick}
      \plotMean{results_flipped_vgg5_backprop_fine_tune_results_accuracy_mean_std.csv}{9.4796}{100}{ForestGreen}{false}{BP domain adaptation}{thick}
      \plotMean{results_flipped_vgg5_pred_code_standalone_results_accuracy_mean_std.csv}{4.3399}{100}{orange}{false}{PC from scratch}{thick}
      \plotMean{results_flipped_vgg5_pred_code_after_backprop_results_accuracy_mean_std.csv}{4.3399}{100}{blue}{false}{PC domain adaptation}{thick}

    \end{axis}
    \end{tikzpicture}
        \end{subfigure}%
        \hspace{-1cm}
        \hfill
        \begin{subfigure}{0.38\textwidth}
            \centering
                \begin{tikzpicture}
    \begin{axis}[
      title={Rotation},
      xlabel style={at={(axis description cs:0.5, -0.1)},anchor=north},
      xlabel={Training time [s]},
      ylabel={Test accuracy (\%)},
        every axis y label/.style={
            at={(ticklabel cs:0.5, 1)},rotate=90,anchor=center,
        },
      grid=major,
      xtick={0, 100, 200, 300},
      width=\textwidth, %
      height=5cm,
      xmax=400,
      ymax=92,
      ymin=48,
      legend pos=south east,
      legend style={legend columns=2,
            /tikz/every even column/.append style={column sep=0.5cm}, at={(0.5,-0.3)}, anchor=north, overlay}
        ]
      \plotMeanAndStdDev{results_rotated_vgg5_backprop_acc_results_accuracy_mean_std.csv}{9.4796}{100}{red}{false}{BP from scratch}{thick}
      \plotMeanAndStdDev{results_rotated_vgg5_backprop_fine_tune_results_accuracy_mean_std.csv}{9.4796}{100}{ForestGreen}{false}{BP domain adaptation}{thick}
      \plotMeanAndStdDev{results_rotated_vgg5_pred_code_standalone_results_accuracy_mean_std.csv}{4.3399}{100}{orange}{false}{PC from scratch}{thick}
      \plotMeanAndStdDev{results_rotated_vgg5_pred_code_after_backprop_results_accuracy_mean_std.csv}{4.3399}{100}{blue}{false}{PC domain adaptation}{thick}

    \end{axis}
    \end{tikzpicture}
        \end{subfigure}%
        \hspace{-1cm}
        \hfill
        \begin{subfigure}{0.38\textwidth}
            \centering
                \begin{tikzpicture}
    \begin{axis}[
      title={Noise},
      xlabel style={at={(axis description cs:0.5, -0.1)},anchor=north},
      xlabel={Training time [s]},
      ylabel={Test accuracy (\%)},
        every axis y label/.style={
            at={(ticklabel cs:0.5, 1)},rotate=90,anchor=center,
        },
      grid=major,
      xtick={0, 100, 200, 300},
      width=\textwidth, %
      height=5cm,
      xmax=400,
      ymax=92,
      ymin=48,
      legend pos=south east,
      legend style={legend columns=2,
            /tikz/every even column/.append style={column sep=0.5cm}},
        ]

      \plotMeanAndStdDev{results_noise_vgg5_backprop_acc_results_accuracy_mean_std.csv}{9.4796}{100}{red}{false}{BP from scratch}{thick}
      \plotMeanAndStdDev{results_noise_vgg5_backprop_fine_tune_results_accuracy_mean_std.csv}{9.4796}{100}{ForestGreen}{false}{BP domain adaptation}{thick}
      \plotMeanAndStdDev{results_noise_vgg5_pred_code_standalone_results_accuracy_mean_std.csv}{4.3399}{100}{orange}{false}{PC from scratch}{thick}
      \plotMeanAndStdDev{results_noise_vgg5_pred_code_after_backprop_results_accuracy_mean_std.csv}{4.3399}{100}{blue}{false}{PC domain adaptation}{thick}

    \end{axis}
    \end{tikzpicture}
        \end{subfigure}
        \hspace{-5mm}
        
        \vspace{3mm}
        \begin{subfigure}{\textwidth}
            \centering
                   \begin{tikzpicture}
          \begin{axis}[
            hide axis,
            width=0.8\textwidth,
            height=2.5cm,
          legend style={legend columns=2,
                /tikz/every even column/.append style={column sep=0.5cm}},
          ]
          \addplot[blue, very thick] coordinates {(0,0) (0,0)};
          \addplot[orange, very thick] coordinates {(0,0) (0,0)};
          \addplot[ForestGreen, very thick] coordinates {(0,0) (0,0)};
          \addplot[red, very thick] coordinates {(0,0) (0,0)};

          \legend{PC Domain Adaptation, PC from scratch, BP Domain Adaptation, BP from scratch}
          \end{axis}
        \end{tikzpicture}
        \end{subfigure}
    \caption{VGG5}
    \label{fig:accuracy_vs_time_vgg5}
    \end{subfigure}
    \begin{subfigure}{\textwidth}
        \centering
        \hspace{-5mm}
        \begin{subfigure}[t]{0.38\textwidth}
            \centering
              \begin{tikzpicture}
    \begin{axis}[
      title={Inversion},
      xlabel style={at={(axis description cs:0.5, -0.1)},anchor=north},
      xlabel={Training time [s]},
      ylabel={Test accuracy (\%)},
      every axis y label/.style={
            at={(ticklabel cs:0.5, 1)},rotate=90,anchor=center,
        },
      grid=major,
      width=\textwidth, %
      height=5cm,
      xmax=450,
      ymax=92,
      ymin=38,
      legend pos=south east,
      legend style={legend columns=2,
            /tikz/every even column/.append style={column sep=0.5cm}}
        ]
       \fillStdDev{results_flipped_vgg7_backprop_acc_results_accuracy_mean_std.csv}{10.4849}{100}{red}
       \fillStdDev{results_flipped_vgg7_backprop_fine_tune_results_accuracy_mean_std.csv}{10.4849}{100}{ForestGreen}
      \fillStdDev{results_flipped_vgg7_pred_code_after_backprop_results_accuracy_mean_std.csv}{5.1576}{100}{blue}
      \fillStdDev{results_flipped_vgg7_pred_code_standalone_results_accuracy_mean_std.csv}{5.1576}{100}{orange}
      
      \plotMean{results_flipped_vgg7_backprop_acc_results_accuracy_mean_std.csv}{10.4849}{100}{red}{false}{BP from scratch}{thick}
      \plotMean{results_flipped_vgg7_backprop_fine_tune_results_accuracy_mean_std.csv}{10.4849}{100}{ForestGreen}{false}{BP domain adaptation}{thick}

      \plotMean{results_flipped_vgg7_pred_code_standalone_results_accuracy_mean_std.csv}{5.1576}{100}{orange}{false}{PC from scratch}{thick}
      \plotMean{results_flipped_vgg7_pred_code_after_backprop_results_accuracy_mean_std.csv}{5.1576}{100}{blue}{false}{PC domain adaptation}{thick}

    \end{axis}
\end{tikzpicture}
         \end{subfigure}%
         \hspace{-1cm}
         \hfill
         \begin{subfigure}{0.38\textwidth}
             \centering
               \begin{tikzpicture}
    \begin{axis}[
      title={Rotation},
      xlabel style={at={(axis description cs:0.5, -0.1)},anchor=north},
      xlabel={Training time [s]},
      ylabel={Test accuracy (\%)},
      every axis y label/.style={
            at={(ticklabel cs:0.5, 1)},rotate=90,anchor=center,
      },
      grid=major,
      width=\textwidth, %
      height=5cm,
      xmax=450,
      ymax=92,
      ymin=38,
      legend pos=south east,
      legend style={legend columns=2,
            /tikz/every even column/.append style={column sep=0.5cm}}
        ]
      \plotMeanAndStdDev{results_rotated_vgg7_backprop_acc_results_accuracy_mean_std.csv}{10.4849}{100}{red}{false}{BP from scratch}{thick}
      \plotMeanAndStdDev{results_rotated_vgg7_backprop_fine_tune_results_accuracy_mean_std.csv}{10.4849}{100}{ForestGreen}{false}{BP domain adaptation}{thick}
      \plotMeanAndStdDev{results_rotated_vgg7_pred_code_standalone_results_accuracy_mean_std.csv}{5.1576}{100}{orange}{false}{PC from scratch}{thick}
      \plotMeanAndStdDev{results_rotated_vgg7_pred_code_after_backprop_results_accuracy_mean_std.csv}{5.1576}{100}{blue}{false}{PC domain adaptation}{thick}
    \end{axis}
\end{tikzpicture}
         \end{subfigure}%
         \hspace{-1cm}
         \hfill
         \begin{subfigure}{0.38\textwidth}
             \centering
               \begin{tikzpicture}
    \begin{axis}[
      title={Noise},
      xlabel style={at={(axis description cs:0.5, -0.1)},anchor=north},
      xlabel={Training time [s]},
      ylabel={Test accuracy (\%)},
      every axis y label/.style={
            at={(ticklabel cs:0.5, 1)},rotate=90,anchor=center,
      },
      grid=major,
      width=\textwidth, %
      height=5cm,
      xmax=450,
      ymax=92,
      ymin=38,
      legend pos=south east,
      legend style={legend columns=2,
            /tikz/every even column/.append style={column sep=0.5cm}}
        ]
      \plotMeanAndStdDev{results_noise_vgg7_backprop_acc_results_accuracy_mean_std.csv}{10.4849}{100}{red}{false}{BP from scratch}{thick}
      \plotMeanAndStdDev{results_noise_vgg7_backprop_fine_tune_results_accuracy_mean_std.csv}{10.4849}{100}{ForestGreen}{false}{BP domain adaptation}{thick}
      \plotMeanAndStdDev{results_noise_vgg7_pred_code_standalone_results_accuracy_mean_std.csv}{5.1576}{100}{orange}{false}{PC from scratch}{thick}
      \plotMeanAndStdDev{results_noise_vgg7_pred_code_after_backprop_results_accuracy_mean_std.csv}{5.1576}{100}{blue}{false}{PC domain adaptation}{thick}
    \end{axis}
\end{tikzpicture}
         \end{subfigure}
         \hspace{-5mm}
    
         \vspace{3mm}
         \begin{subfigure}{\textwidth}
            \centering
                   \begin{tikzpicture}
          \begin{axis}[
            hide axis,
            width=0.8\textwidth,
            height=2.5cm,
          legend style={legend columns=2,
                /tikz/every even column/.append style={column sep=0.5cm}},
          ]
          \addplot[blue, very thick] coordinates {(0,0) (0,0)};
          \addplot[orange, very thick] coordinates {(0,0) (0,0)};
          \addplot[ForestGreen, very thick] coordinates {(0,0) (0,0)};
          \addplot[red, very thick] coordinates {(0,0) (0,0)};

          \legend{PC Domain Adaptation, PC from scratch, BP Domain Adaptation, BP from scratch}
          \end{axis}
        \end{tikzpicture}
         \end{subfigure}
     \caption{VGG7}
    \label{fig:accuracy_vs_time_vgg7}
    \end{subfigure}
\end{figure}
\begin{figure}[!t]\ContinuedFloat
     \begin{subfigure}{\textwidth}    
        \centering
        \hspace{-5mm}
        \begin{subfigure}[t]{0.38\textwidth}
            \centering
                \begin{tikzpicture}
    \begin{axis}[
      title={Inversion},
      xlabel style={at={(axis description cs:0.5, -0.1)},anchor=north},
      xlabel={Training time [s]},
      ylabel={Test accuracy (\%)},
      every axis y label/.style={
            at={(ticklabel cs:0.5, 1)},rotate=90,anchor=center,
        },
      grid=major,
      width=\textwidth, %
      height=5cm,
      xmax=450,
      ymax=95,
      ymin=5,
      legend pos=south east,
      legend style={legend columns=2,
            /tikz/every even column/.append style={column sep=0.5cm}}
        ]
      \fillStdDev{results_flipped_vgg9_backprop_acc_results_accuracy_mean_std.csv}{8.4814}{100}{red}
      \fillStdDev{results_flipped_vgg9_backprop_fine_tune_results_accuracy_mean_std.csv}{8.4814}{100}{ForestGreen}
      \fillStdDev{results_flipped_vgg9_pred_code_after_backprop_results_accuracy_mean_std.csv}{5.4290}{100}{blue}
      \fillStdDev{results_flipped_vgg9_pred_code_standalone_results_accuracy_mean_std.csv}{5.4290}{100}{orange}

      \plotMean{results_flipped_vgg9_backprop_acc_results_accuracy_mean_std.csv}{8.4814}{100}{red}{false}{BP from scratch}{thick}
      \plotMean{results_flipped_vgg9_backprop_fine_tune_results_accuracy_mean_std.csv}{8.4814}{100}{ForestGreen}{false}{BP domain adaptation}{thick}

      \plotMean{results_flipped_vgg9_pred_code_standalone_results_accuracy_mean_std.csv}{5.4290}{100}{orange}{false}{PC from scratch}
      {thick}\plotMean{results_flipped_vgg9_pred_code_after_backprop_results_accuracy_mean_std.csv}{5.4290}{100}{blue}{false}{PC domain adaptation}{thick}
    \end{axis}
    \end{tikzpicture}
         \end{subfigure}%
         \hspace{-1cm}
         \hfill
         \begin{subfigure}{0.38\textwidth}
             \centering
                 \begin{tikzpicture}
    \begin{axis}[
      title={Rotation},
      xlabel style={at={(axis description cs:0.5, -0.1)},anchor=north},
      xlabel={Training time [s]},
      ylabel={Test accuracy (\%)},
      every axis y label/.style={
            at={(ticklabel cs:0.5, 1)},rotate=90,anchor=center,
        },
      grid=major,
      width=\textwidth, %
      height=5cm,
      xmax=450,
      ymax=95,
      ymin=5,
      legend pos=south east,
      legend style={legend columns=2,
            /tikz/every even column/.append style={column sep=0.5cm}}
        ]
      \plotMeanAndStdDev{results_rotated_vgg9_backprop_acc_results_accuracy_mean_std.csv}{8.4814}{100}{red}{false}{BP from scratch}{thick}
      \plotMeanAndStdDev{results_rotated_vgg9_backprop_fine_tune_results_accuracy_mean_std.csv}{8.4814}{100}{ForestGreen}{false}{BP domain adaptation}{thick}
      \plotMeanAndStdDev{results_rotated_vgg9_pred_code_standalone_results_accuracy_mean_std.csv}{5.4290}{100}{orange}{false}{PC from scratch}
      {thick}
      \plotMeanAndStdDev{results_rotated_vgg9_pred_code_after_backprop_results_accuracy_mean_std.csv}{5.4290}{100}{blue}{false}{PC domain adaptation}{thick}
    \end{axis}
    \end{tikzpicture}
         \end{subfigure}%
         \hspace{-1cm}
         \hfill
         \begin{subfigure}{0.38\textwidth}
             \centering
                 \begin{tikzpicture}
    \begin{axis}[
      title={Noise},
      xlabel style={at={(axis description cs:0.5, -0.1)},anchor=north},
      xlabel={Training time [s]},
      ylabel={Test accuracy (\%)},
      every axis y label/.style={
            at={(ticklabel cs:0.5, 1)},rotate=90,anchor=center,
        },
      grid=major,
      width=\textwidth, %
      height=5cm,
      xmax=450,
      ymax=95,
      ymin=5,
      legend pos=south east,
      legend style={legend columns=2,
            /tikz/every even column/.append style={column sep=0.5cm}}
        ]
      \plotMeanAndStdDev{results_noise_vgg9_backprop_acc_results_accuracy_mean_std.csv}{8.4814}{100}{red}{false}{BP from scratch}{thick}
      \plotMeanAndStdDev{results_noise_vgg9_backprop_fine_tune_results_accuracy_mean_std.csv}{8.4814}{100}{ForestGreen}{false}{BP domain adaptation}{thick}
      \plotMeanAndStdDev{results_noise_vgg9_pred_code_standalone_results_accuracy_mean_std.csv}{5.4290}{100}{orange}{false}{PC from scratch}
      {thick}
      \plotMeanAndStdDev{results_noise_vgg9_pred_code_after_backprop_results_accuracy_mean_std.csv}{5.4290}{100}{blue}{false}{PC domain adaptation}{thick}
    \end{axis}
    \end{tikzpicture}
         \end{subfigure}
         \hspace{-5mm}
    
         \vspace{3mm}
         \begin{subfigure}{\textwidth}
            \centering
                   \begin{tikzpicture}
          \begin{axis}[
            hide axis,
            width=0.8\textwidth,
            height=2.5cm,
          legend style={legend columns=2,
                /tikz/every even column/.append style={column sep=0.5cm}},
          ]
          \addplot[blue, very thick] coordinates {(0,0) (0,0)};
          \addplot[orange, very thick] coordinates {(0,0) (0,0)};
          \addplot[ForestGreen, very thick] coordinates {(0,0) (0,0)};
          \addplot[red, very thick] coordinates {(0,0) (0,0)};

          \legend{PC Domain Adaptation, PC from scratch, BP Domain Adaptation, BP from scratch}
          \end{axis}
        \end{tikzpicture}
         \end{subfigure}
     \caption{VGG9}
     \label{fig:accuracy_vs_time_vgg9}
     \end{subfigure}
    \begin{subfigure}{\textwidth}    
        \centering
        \hspace{-5mm}
        \begin{subfigure}[t]{0.38\textwidth}
            \centering
              \begin{tikzpicture}
    \begin{axis}[
      title={Inversion},
      xlabel style={at={(axis description cs:0.5, -0.1)},anchor=north},
      xlabel={Training time [s]},
      ylabel={Test accuracy (\%)},
      every axis y label/.style={
          at={(ticklabel cs:0.5, 1)},rotate=90,anchor=center,
      },
      grid=major,
      width=\textwidth, %
      height=5cm,
      xmax=19,
      legend pos=south east,
      legend style={legend columns=2,
            /tikz/every even column/.append style={column sep=0.5cm}},
        ]      
      \plotMeanAndStdDev{results_flipped_mlp_backprop_acc_results_domain_shifted_accuracy_mean_std.csv}{1.8820}{100}{red}{false}{BP from scratch}{thick}
      \plotMeanAndStdDev{results_flipped_mlp_backprop_fine_tune_results_accuracy_mean_std.csv}{1.8820}{100}{ForestGreen}{false}{BP domain adaptation}{thick}
      \plotMeanAndStdDev{results_flipped_mlp_pred_code_standalone_results_accuracy_mean_std.csv}{2.9256}{100}{orange}{false}{PC from scratch}{thick}
      \plotMeanAndStdDev{results_flipped_mlp_pred_code_after_backprop_results_accuracy_mean_std.csv}{2.9256}{100}{blue}{false}{PC domain adaptation}{thick}
    \end{axis}
\end{tikzpicture}
         \end{subfigure}%
         \hspace{-1cm}
         \hfill
         \begin{subfigure}{0.38\textwidth}
             \centering
               \begin{tikzpicture}
    \begin{axis}[
      title={Rotation},
      xlabel style={at={(axis description cs:0.5, -0.1)},anchor=north},
      xlabel={Training time [s]},
      ylabel={Test accuracy (\%)},
      every axis y label/.style={
          at={(ticklabel cs:0.5, 1)},rotate=90,anchor=center,
      },
      grid=major,
      width=\textwidth, %
      height=5cm,
      xmax=19,
      legend pos=south east,
      legend style={legend columns=2,
            /tikz/every even column/.append style={column sep=0.5cm}},
        ]      
      \plotMeanAndStdDev{results_rotated_mlp_backprop_acc_results_domain_shifted_accuracy_mean_std.csv}{1.8820}{100}{red}{false}{BP from scratch}{thick}
      \plotMeanAndStdDev{results_rotated_mlp_backprop_fine_tune_results_accuracy_mean_std.csv}{1.8820}{100}{ForestGreen}{false}{BP domain adaptation}{thick}
      \plotMeanAndStdDev{results_rotated_mlp_pred_code_standalone_results_accuracy_mean_std.csv}{2.9256}{100}{orange}{false}{PC from scratch}{thick}
      \plotMeanAndStdDev{results_rotated_mlp_pred_code_after_backprop_results_accuracy_mean_std.csv}{2.9256}{100}{blue}{false}{PC domain adaptation}{thick}
    \end{axis}
\end{tikzpicture}
         \end{subfigure}%
         \hspace{-1cm}
         \hfill
         \begin{subfigure}{0.38\textwidth}
             \centering
               \begin{tikzpicture}
    \begin{axis}[
      title={Noise},
      xlabel style={at={(axis description cs:0.5, -0.1)},anchor=north},
      xlabel={Training time [s]},
      ylabel={Test accuracy (\%)},
      every axis y label/.style={
          at={(ticklabel cs:0.5, 1)},rotate=90,anchor=center,
      },
      grid=major,
      width=\textwidth, %
      height=5cm,
      xmax=19,
      legend pos=south east,
      legend style={legend columns=2,
            /tikz/every even column/.append style={column sep=0.5cm}},
        ]      
      \plotMeanAndStdDev{results_noise_mlp_backprop_acc_results_domain_shifted_accuracy_mean_std.csv}{1.8820}{100}{red}{false}{BP from scratch}{thick}
      \plotMeanAndStdDev{results_noise_mlp_backprop_fine_tune_results_accuracy_mean_std.csv}{1.8820}{100}{ForestGreen}{false}{BP domain adaptation}{thick}
      \plotMeanAndStdDev{results_noise_mlp_pred_code_standalone_results_accuracy_mean_std.csv}{2.9256}{100}{orange}{false}{PC from scratch}{thick}
      \plotMeanAndStdDev{results_noise_mlp_pred_code_after_backprop_results_accuracy_mean_std.csv}{2.9256}{100}{blue}{false}{PC domain adaptation}{thick}
    \end{axis}
\end{tikzpicture}
         \end{subfigure}
         \hspace{-5mm}
    
         \vspace{3mm}
         \begin{subfigure}{\textwidth}
            \centering
                   \begin{tikzpicture}
          \begin{axis}[
            hide axis,
            width=0.8\textwidth,
            height=2.5cm,
          legend style={legend columns=2,
                /tikz/every even column/.append style={column sep=0.5cm}},
          ]
          \addplot[blue, very thick] coordinates {(0,0) (0,0)};
          \addplot[orange, very thick] coordinates {(0,0) (0,0)};
          \addplot[ForestGreen, very thick] coordinates {(0,0) (0,0)};
          \addplot[red, very thick] coordinates {(0,0) (0,0)};

          \legend{PC Domain Adaptation, PC from scratch, BP Domain Adaptation, BP from scratch}
          \end{axis}
        \end{tikzpicture}
         \end{subfigure}
         \caption{MLP}
     \label{fig:accuracy_vs_time_mlp}
     \end{subfigure}
     \caption{Model Accuracy as a function of training time for models trained from scratch and domain-adapted using Backpropagation (BP) and Predictive Coding (PC). Trainings were performed, for each model, respectively with inverted, rotated and noisy data (Section \ref{subsec:experimental_setup}). The plots highlight the validity brought by the proposed technique, especially for the VGG5 model. The transparent areas represent the accuracy standard deviation over five runs.}
     \label{fig:accuracy_vs_time}
\end{figure}
The performance of PC-based algorithms decreases as model depth increases, as is evident for the deeper VGG9 model. This aligns with previous observations that PC struggles when training deep architectures \cite{goemaere2025erroroptimizationovercomingexponential, pinchetti2025benchmarkingpredictivecodingnetworks}, while it can outperform BP for more shallow architectures \cite{pinchetti2025benchmarkingpredictivecodingnetworks}.
The BP-based domain adaptation for VGG9 delivers lower performances with respect to the BP-based domain adaptation of the other models, highlighting that domain adaptation may in general be more difficult for deeper models,  regardless of the training technique.
For the easier MNIST dataset with a simple three layer MLP model, both the PC and BP domain adaptation prove to be beneficial, providing equally accurate models. Both approaches outperform training a model from scratch. 

\FloatBarrier
\section{Conclusion and future work}
\label{sec:conclusion}
Our experiments demonstrate that Predictive Coding (PC) is an efficient algorithm for on-device domain adaptation of models initially pretrained with Backpropagation (BP). We showed that models trained with BP can effectively leverage PC-based finetuning on a domain-shifted variant of the same dataset—in our case, MNIST and CIFAR-10. This opens up new avenues of research into computationally efficient domain adaptation techniques using PC. This could prove especially efficient when implemented on neuromorphic hardware \cite{ajan2022advances_in_neuromorphic_devices}.

Future work will extend this study by evaluating training times on embedded platforms, including neuromorphic hardware. We also aim to scale our approach to deeper neural network architectures. While the domain adaptation procedure was completely supervised in this work, this might not be feasible in real-world environments. More research into unsupervised or self-supervised approaches in combination with PC is therefore a highly interesting research direction.

\section*{Acknowledgments}
This research was supported by funding from the Flemish Government under the ``Onderzoeksprogramma Artifici\"ele Intelligentie (AI) Vlaanderen'' program.

\bibliography{references}
\bibliographystyle{unsrt}

\newpage
\section*{Appendix}

Table \ref{tab:mlp} reports the MLP architecture that was used in this study, while Table \ref{tab:vgg_models} describes the VGG-inspired architectures \cite{simonyan2015deepconvolutionalnetworkslargescale}. Both the MLP and VGG architectures follow the setup from \cite{pinchetti2025benchmarkingpredictivecodingnetworks}.

\begin{table}[h!]
\centering
\caption{Summary of the MLP model for MNIST classification}
\vspace{3mm}
    \begin{tabular}{l}
         \toprule
         \textbf{MLP}\\
         \midrule
         \makecell[l]{Dense(784, 512)\\Leaky ReLU}\\\hline
         \makecell[l]{Dense(512, 512)\\Leaky ReLU}\\\hline
         Dense(512, 10)\\\bottomrule
    \end{tabular}
    \label{tab:mlp}
\end{table}

\begin{table}[H]
\centering
\caption{Summary of The VGG5, VGG7, and VGG9 models, inspired from \cite{simonyan2015deepconvolutionalnetworkslargescale}. The notation Conv<$n$>\textunderscore<$m$> stands for a 2D convolutional layer with a kernel of size $n\times n$ and $m$ channels. All Convolutional layers are followed by a GELU non-linear activation. All Max Pooling layers have a size of $2 \times 2$ and a stride of $2 \times 2$.}
\label{tab:vgg_models}
\begin{tabularx}{\textwidth}{>{\raggedright\arraybackslash}X >{\raggedright\arraybackslash}X >{\raggedright\arraybackslash}X}
    \toprule
    \textbf{VGG5} & \textbf{VGG7} & \textbf{VGG9}\\ 
    \midrule
    \makecell[l]{Conv3-128 (Pad (1, 1))\\Max Pool} & \makecell[l]{Conv3-128 (Pad (1, 1))\\Max Pool} & \makecell[l]{Conv3-128 (Pad (1, 1))\\Max Pool}\\\midrule
    \makecell[l]{Conv3-256 (Pad (1, 1))\\Max Pool} & \makecell[l]{Conv3-128 (Pad (1, 1))} & \makecell[l]{Conv3-128 (Pad (1, 1))\\Max Pool}\\\midrule
    \makecell[l]{Conv3-512 (Pad (1, 1))\\Max Pool} & \makecell[l]{Conv3-256 (Pad (1, 1))} & \makecell[l]{Conv3-256  (Pad (1, 1))\\Max Pool}\\\midrule
    \makecell[l]{Conv3-512 (Pad (1, 1))\\Max Pool} & \makecell[l]{Conv3-256} & \makecell[l]{Conv3-256  (Pad (1, 1))\\Max Pool}\\\midrule
    - & \makecell[l]{Conv3-512 (Pad (1, 1))\\Max Pool} & \makecell[l]{Conv3-512 (Pad (1, 1))}\\\hline
    - & \makecell[l]{Conv3-512} & \makecell[l]{Conv3-512 (Pad (1, 1))\\Max Pool}\\\hline
    - & - & \makecell[l]{Conv3-512 (Pad (1, 1))}\\\midrule
    - & - & \makecell[l]{Conv3-512 (Pad (1, 1))}\\\midrule
    Dense(2048, 10) & Dense(512,10) & Dense(512, 10)\\\bottomrule
\end{tabularx}
\end{table}

\noindent Tables \ref{tab:training_hyperparams_flipped}, \ref{tab:training_hyperparams_rotated}, and \ref{tab:training_hyperparams_noise} describe the hyperparameters that were used to obtain the results from Figure \ref{fig:accuracy_vs_time}.\\
Using the design space search from \cite{pinchetti2025benchmarkingpredictivecodingnetworks} as a starting point:
\begin{itemize}
    \item Both Squared Error (SE) and Cross Entropy (CE) were used as loss function, also in Predictive Coding(PC)-based training. In fact, the energy function can be adapted, as explained in \cite{pinchetti2022predictive_coding_beyond_data_distribution}.
    \item AdamW (Adam with weight decay regularization) was used as weights optimizer.
    \item SGD with momentum was used as predictions optimizer (for the PC-based trainings).
    \item All the weights learning rate was scheduled with a linear warmup followed by cosine decay.
\end{itemize}
\noindent However, unlike in \cite{pinchetti2025benchmarkingpredictivecodingnetworks}, the number of PC inference steps (Subsection \ref{subsec:predictive_coding}) was kept at 4 for all CNN models. This choice was taken to obtain a training time per epoch that is, for each model, approximately half the time per epoch of the Backpropagation (BP) training counterpart. All PC trainings were performed with Forward Initialization, which is widely used to accelerate the predictions optimization \cite{goemaere2025erroroptimizationovercomingexponential}. This initialization technique implies that for each training batch a first forward pass (BP-like) is performed, as in Equation \ref{eq_forward_pass_layer_operation}. After the forward pass, the predictions $\mu_l$ for are initialized as the activities $a_l$, for each layer apart from the last one, for which the labels $y$ are used. After this step, the training takes place as in Subsection \ref{subsec:predictive_coding}.

\begin{table}
    \caption{Models training hyperparameters corresponding to each specific training technique: training from scratch and domain adaptation performed both with Backpropagation (BP) and Predictive Coding (PC). The training was performed on data subject to different domain shifts: inversion, rotation, and noise addition (see Subsection \ref{subsec:experimental_setup} for details). AdamW was used as weights optimizer, while SGD with momentum was used as predictions optimizer.}
    \label{tab:training_hyperparams}
    \begin{subtable}{\textwidth}
        \caption{Inverted images}
        \label{tab:training_hyperparams_flipped}
            \begin{threeparttable}
    \centering
    \footnotesize
    \begin{tabularx}{\textwidth}{p{2cm} >{\raggedright\arraybackslash}X p{1cm}  >{\raggedright\arraybackslash}X >{\raggedright\arraybackslash}X  >{\raggedright\arraybackslash}X >{\raggedright\arraybackslash}X >{\raggedright\arraybackslash}X >{\raggedright\arraybackslash}X}
        \toprule
        & \bf Model & \bf Loss & $\mathbf{\theta_{lr}}$\tnote{*} & \bf Weight decay & $\mathbf{\gamma}$\tnote{\dag} & $\mathbf{m}$\tnote{\ddag} & \bf Inference steps\\
        \midrule
        \multirow{4}{*}{\makecell[l]{BP\\from\\ scratch}} & VGG5 & SE & $2.5e^{-4}$ & $3e^{-4}$ & \multirow{4}{*}{-} & \multirow{4}{*}{-} & \multirow{4}{*}{-}\\
                                         & VGG7 & SE & $2.5e^{-4}$ & $2e^{-4}$ & & & \\
                                         & VGG9 & CE & $5e^{-4}$ & $3e^{-4}$ & & & \\
                                         & MLP & CE & $1e^{-3}$ & $1e^{-4}$ & & & \\\midrule
        \multirow{4}{*}{\makecell[l]{BP\\domain\\ adaptation}} & VGG5 & SE & $1e^{-3}$ & $2e^{-4}$ & \multirow{4}{*}{-} & \multirow{4}{*}{-} & \multirow{4}{*}{-}\\
                                         & VGG7 & CE & $1e^{-3}$ & $2e^{-4}$ & & & \\
                                         & VGG9 & CE & $1e^{-3}$ & $3e^{-4}$ & & & \\
                                         & MLP & CE & $1e^{-3}$ & $1e^{-4}$ & & & \\\midrule
        \multirow{4}{*}{\makecell[l]{PC\\from\\ scratch}}  & VGG5 & SE & $1e^{-4}$ & $1e^{-4}$ & $2.5e^{-2}$ & 0.1 & 4\\
                                          & VGG7 & CE & $1e^{-4}$ & $1e^{-4}$ & $2.5e^{-2}$ & 0.1 & 4 \\
                                          & VGG9 & SE & $1e^{-3}$ & $1e^{-4}$ & $1e^{-4}$ & 0.1 & 4 \\
                                          & MLP & CE & $1e^{-3}$ & $1e^{-4}$ & $1e^{-3}$ & 0.1 & 13\\\midrule
        \multirow{4}{*}{\makecell[l]{PC\\domain\\adaptation}} & VGG5 & CE & $1e^{-4}$ & $1e^{-4}$ & $1e^{-2}$ & 0.1 & 4\\
                                          & VGG7 & CE & $1e^{-4}$ & $1e^{-4}$ & $1e^{-2}$ & 0.5 & 4\\
                                          & VGG9 & CE & $1e^{-4}$ & $1e^{-4}$ & $1e^{-3}$ & 0.1 & 4\\
                                         & MLP  & CE & $1e^{-3}$ & $1e^{-4}$ & $1e^{-3}$ & 0.1 & 13\\
        \bottomrule
    \end{tabularx}
    \begin{tablenotes}
       \item [*] Weights learning rate.
       \item [\dag] Inference rate.
       \item [\ddag] SGD optimizer momentum.
    \end{tablenotes}
\end{threeparttable}

    \end{subtable}
\end{table}
\begin{table}\ContinuedFloat
    \begin{subtable}{\textwidth}
        \caption{Rotated images}
        \label{tab:training_hyperparams_rotated}
            \begin{threeparttable}
    \centering
    \footnotesize
    \begin{tabularx}{\textwidth}{p{2cm} >{\raggedright\arraybackslash}X p{1cm}  >{\raggedright\arraybackslash}X >{\raggedright\arraybackslash}X  >{\raggedright\arraybackslash}X >{\raggedright\arraybackslash}X >{\raggedright\arraybackslash}X >{\raggedright\arraybackslash}X}
        \toprule
        & \bf Model & \bf Loss & $\mathbf{\theta_{lr}}$\tnote{*} & \bf Weight decay & $\mathbf{\gamma}$\tnote{\dag} & $\mathbf{m}$\tnote{\ddag} & \bf Inference steps\\
        \midrule
        \multirow{4}{*}{\makecell[l]{BP\\from\\ scratch}} & VGG5 & SE & $2.5e^{-4}$ & $2e^{-4}$ & \multirow{4}{*}{-} & \multirow{4}{*}{-} & \multirow{4}{*}{-}\\
         & VGG7 & SE & $2.5e^{-4}$ & $3e^{-4}$ & & & \\
         & VGG9 & SE & $5e^{-4}$ & $2e^{-4}$ & & & \\
         & MLP & CE & $1e^{-3}$ & $1e^{-4}$ & & & \\\midrule
        \multirow{4}{*}{\makecell[l]{BP\\domain\\adaptation}} & VGG5 & SE & $5e^{-4}$ & $1e^{-4}$ & \multirow{4}{*}{-} & \multirow{4}{*}{-} & \multirow{4}{*}{-}\\
         & VGG7 & CE & $1e^{-3}$ & $1e^{-4}$ & & & \\
         & VGG9 & SE & $1e^{-3}$ & $3e^{-4}$ & & & \\
         & MLP & CE & $1e^{-3}$ & $1e^{-4}$ & & & \\\midrule
        \multirow{4}{*}{\makecell[l]{PC\\from\\scratch}} & VGG5 & SE & $1e^{-4}$ & $1e^{-4}$ & $2.5e^{-2}$ & 0.5 & 4 \\
         & VGG7 & CE & $1e^{-4}$ & $1e^{-4}$ & $2.5e^{-2}$ & 0.1 & 4 \\
         & VGG9 & SE & $1e^{-4}$ & $1e^{-4}$ & $1e^{-2}$ & 0.1 & 4 \\
         & MLP  & CE & $1e^{-3}$ & $1e^{-4}$ & $1e^{-3}$ & 0.1 & 13\\\midrule
        \multirow{4}{*}{\makecell[l]{PC\\domain\\adaptation}} & VGG5 & CE & $1e^{-4}$ & $1e^{-4}$ & $1e^{-2}$ & 0.1 & 4 \\
         & VGG7 & CE & $1e^{-4}$ & $1e^{-4}$ & $1e^{-2}$ & 0.1 & 4 \\
         & VGG9 & CE & $1e^{-4}$ & $1e^{-4}$ & $1e^{-2}$ & 0.1 & 4 \\
         & MLP  & CE & $1e^{-3}$ & $1e^{-4}$ & $1e^{-3}$ & 0.1 & 13\\
\bottomrule
    \end{tabularx}
    \begin{tablenotes}
       \item [*] Weights learning rate.
       \item [\dag] Inference rate.
       \item [\ddag] SGD optimizer momentum.
    \end{tablenotes}
\end{threeparttable}

    \end{subtable}
\end{table}
\begin{table}\ContinuedFloat
    \begin{subtable}{\textwidth}
        \caption{Noisy images}
        \label{tab:training_hyperparams_noise}
            \begin{threeparttable}
    \centering
    \footnotesize
    \begin{tabularx}{\textwidth}{p{2cm} >{\raggedright\arraybackslash}X p{1cm}  >{\raggedright\arraybackslash}X >{\raggedright\arraybackslash}X  >{\raggedright\arraybackslash}X >{\raggedright\arraybackslash}X >{\raggedright\arraybackslash}X >{\raggedright\arraybackslash}X}
        \toprule
        & \bf Model & \bf Loss & $\mathbf{\theta_{lr}}$\tnote{*} & \bf Weight decay & $\mathbf{\gamma}$\tnote{\dag} & $\mathbf{m}$\tnote{\ddag} & \bf Inference steps\\
        \midrule
        \multirow{4}{*}{\makecell[l]{BP\\from\\ scratch}} & VGG5 & SE & $2.5e^{-4}$ & $1e^{-4}$ & \multirow{4}{*}{-} & \multirow{4}{*}{-} & \multirow{4}{*}{-}\\
         & VGG7 & SE & $2.5e^{-4}$ & $1e^{-4}$ & & & \\
         & VGG9 & SE & $2.5e^{-4}$ & $1e^{-4}$ & & & \\
         & MLP & CE & $1e^{-3}$ & $1e^{-4}$ & & & \\\midrule
        \multirow{4}{*}{\makecell[l]{BP\\domain\\adaptation}} & VGG5 & CE & $2.5e^{-4}$ & $3e^{-4}$ & \multirow{4}{*}{-} & \multirow{4}{*}{-} & \multirow{4}{*}{-}\\
         & VGG7 & CE & $1e^{-4}$ & $2e^{-4}$ & & & \\
         & VGG9 & SE & $5e^{-4}$ & $1e^{-4}$ & & & \\
         & MLP & CE & $1e^{-3}$ & $1e^{-4}$ & & & \\\midrule
        \multirow{4}{*}{\makecell[l]{PC\\from\\scratch}} & VGG5 & SE & $1e^{-4}$ & $1e^{-4}$ & $2.5e^{-2}$ & 0.5 & 4 \\
         & VGG7 & CE & $1e^{-4}$ & $1e^{-4}$ & $2.5e^{-2}$ & 0.1 & 4 \\
         & VGG9 & SE & $1e^{-4}$ & $1e^{-4}$ & $1e^{-2}$ & 0.1 & 4 \\
         & MLP  & CE & $1e^{-3}$ & $1e^{-4}$ & $1e^{-3}$ & 0.1 & 13\\\midrule
        \multirow{4}{*}{\makecell[l]{PC\\domain\\adaptation}} & VGG5 & CE & $1e^{-4}$ & $1e^{-4}$ & $1e^{-2}$ & 0.5 & 4 \\
         & VGG7 & CE & $1e^{-4}$ & $1e^{-4}$ & $2.5e^{-2}$ & 0.5 & 4 \\
         & VGG9 & CE & $1e^{-4}$ & $1e^{-4}$ & $1e^{-2}$ & 0.1 & 4 \\
         & MLP  & CE & $1e^{-3}$ & $1e^{-4}$ & $1e^{-3}$ & 0.1 & 13\\
        \bottomrule
    \end{tabularx}
    \begin{tablenotes}
       \item [*] Weights learning rate.
       \item [\dag] Inference rate.
       \item [\ddag] SGD optimizer momentum.
    \end{tablenotes}
\end{threeparttable}

    \end{subtable}
\end{table}

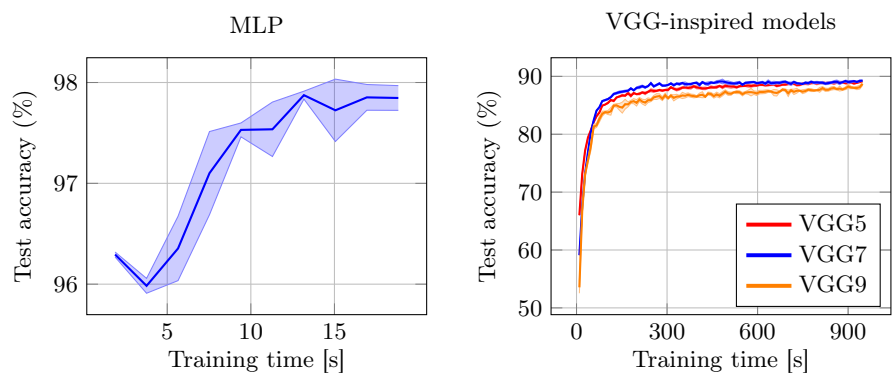
\begin{figure}
    \begin{subfigure}{0.5\textwidth}
        \centering
            \begin{tikzpicture}
    \begin{axis}[
      title={MLP},
      xlabel style={at={(axis description cs:0.5, -0.1)},anchor=north},
      xlabel={Training time [s]},
      ylabel={Test accuracy (\%)},
      grid=major,
      width=\textwidth, %
      height=5cm,
      xtick={0, 5, 10, 15},
      legend pos=south east,
      legend style={legend columns=3,
            /tikz/every even column/.append style={column sep=0.5cm}},
        ]
      \plotMeanAndStdDev{results_original_mlp_backprop_acc_results_accuracy_mean_std.csv}{1.8820}{100}{blue}{false}{MLP}{thick}

    \end{axis}
    \end{tikzpicture}
    \end{subfigure}
    \hfill
    \begin{subfigure}{0.5\textwidth}
        \centering
            \begin{tikzpicture}
    \begin{axis}[
      title={VGG-inspired models},
      xlabel style={at={(axis description cs:0.5, -0.1)},anchor=north},
      xlabel={Training time [s]},
      ylabel={Test accuracy (\%)},
      grid=major,
      width=\textwidth, %
      height=5cm,
      xtick={0, 300, 600, 900},
      legend pos=south east,
      legend style={legend columns=1,
            /tikz/every even column/.append style={column sep=0.5cm}},
        ]
      \plotMeanAndStdDev{results_original_vgg5_backprop_acc_results_accuracy_mean_std.csv}{9.4796}{100}{red}{true}{VGG5}{thick}
      \plotMeanAndStdDev{results_original_vgg7_backprop_acc_results_accuracy_mean_std.csv}{9.4796}{100}{blue}{true}{VGG7}{thick}
      \plotMeanAndStdDev{results_original_vgg9_backprop_acc_results_accuracy_mean_std.csv}{9.4796}{100}{orange}{true}{VGG9}{thick}

    \end{axis}
    \end{tikzpicture}
    \end{subfigure}
    \caption{Test accuracy per training time of the MLP (Table \ref{tab:mlp}) and the VGG-inspired  models (Table \ref{tab:vgg_models}), training with Backpropagation (BP) on original data.}
    \label{fig:accuracy_vs_time_original_data}
\end{figure}

\noindent Figure \ref{fig:accuracy_vs_time_original_data} displays the test accuracy per training time, obtained with the models from Table \ref{tab:mlp} and Table \ref{tab:vgg_models} respectively, training them with BP on original data (i.e. not subject to domain shift). The weights obtained from this models were used as starting point for both the PC and BP-based domain adaptation techniques. 

\begin{table}[H]
    \begin{threeparttable}
    \centering
    \caption{Models Backpropagation (BP) training hyperparameters, used for training from scratch on original data. AdamW was used as weights optimizer.}
    \label{tab:training_hyperparams_original_data}
    \footnotesize
    \begin{tabularx}{\textwidth}{>{\raggedright\arraybackslash}X >{\raggedright\arraybackslash}X >{\raggedright\arraybackslash}X  >{\raggedright\arraybackslash}X}
        \toprule
        \bf Model & \bf Loss & $\mathbf{\theta_{lr}}$\tnote{*} & \bf Weight decay\\
        \midrule
        VGG5 & SE & $2.5e^{-4}$ & $3e^{-4}$\\
        VGG7 & SE & $2.5e^{-4}$ & $2e^{-4}$\\
        VGG9 & CE & $5e^{-4}$ & $3e^{-4}$\\
        MLP  & CE & $1e^{-3}$ & $1e^{-4}$\\
        \bottomrule
    \end{tabularx}
    \begin{tablenotes}
       \item [*] Weights learning rate.
    \end{tablenotes}
\end{threeparttable}
\end{table}

\noindent Table \ref{tab:training_hyperparams_original_data} reports the hyperparameters that were used to train the models with BP on original data (Figure \ref{fig:accuracy_vs_time_original_data}). 

\noindent Table \ref{tab:train_hyperparam_search} describes the CNNs training hyperparameters in which the design space exploration took place.

\begin{table}[h!]
    \caption{Training hyperparameters used for the search of the best model for the Backpropagation (BP) and the Predictive Coding (PC) training.}
    \label{tab:train_hyperparam_search}
    \centering
    \begin{threeparttable}
    \begin{tabularx}{\textwidth}{>{\raggedright\arraybackslash}X p{1.2cm} >{\raggedright\arraybackslash}X p{2cm} >{\raggedright\arraybackslash}X >{\raggedright\arraybackslash}X >{\raggedright\arraybackslash}X >{\raggedright\arraybackslash}X}
        \toprule
         & & & \multicolumn{3}{c}{\bf PC}\\\cmidrule{4-6}
         & & \bf \makecell[l]{BP\\ VGG5-7-9} & \bf VGG5 & \bf VGG7 & \bf VGG9 \\\midrule
        & Loss & [SE, CE] & [SE, CE] & [SE, CE] & [SE, CE]\\\midrule
        \multirow{4}{*}{\makecell[l]{Weights\\update}} & \multirow{2}{*}{$\theta_{lr}$\tnote{*}} & [$1e^{-4}$, $2.5e^{-4}$, $5e^{-4}$, $1e^{-3}$\tnote{\S}] & \makecell[l]{[$1e^{-4}$,\\ $1e^{-3}$]} & \makecell[l]{[$1e^{-4}$,\\ $1e^{-3}$]} & \makecell[l]{[$1e^{-4}$,\\ $1e^{-3}$]} \\\cmidrule{2-6}
        & \makecell[l]{Weight\\decay} & [1, 2, 3]$e^{-4}$ & 1$e^{-4}$ & 1$e^{-4}$ & 1$e^{-4}$\\\midrule
        \multirow{5}{*}{\makecell[l]{Predictions\\update}} & $\gamma$\tnote{\dag} & \makecell{-} & \makecell[l]{[$1e^{-5}$,\\ $1e^{-4}$,\\ $1e^{-3}$,\\ $1e^{-2}$,\\ $2.5e^{-2}$,\\ $5e^{-2}$]} & \makecell[l]{[$1e^{-5}$,\\ $1e^{-4}$,\\ $1e^{-3}$,\\ $1e^{-2}$,\\ $2.5e^{-2}$,\\ $5e^{-2}$]} & \makecell[l]{[$1e^{-5}$,\\ $1e^{-4}$,\\ $1e^{-3}$,\\ $1e^{-2}$]} \\\cmidrule{2-6}
        & m\tnote{\ddag} & \makecell{-} & \makecell[l]{[0.1,\\ 0.5]} & \makecell[l]{[0.1,\\ 0.5]} & \makecell[l]{[0.1,\\ 0.5,\\ 0.9]} \\
        \bottomrule
    \end{tabularx}
        \begin{tablenotes}
       \item [*] Weights learning rate.
       \item [\dag] Inference rate.
       \item [\ddag] SGD optimizer momentum.
       \item [\S] A weights learning rate ($\theta_{lr}$) of 1$e^{-3}$ was used only for BP domain adaptation.
    \end{tablenotes}
    \end{threeparttable}
\end{table}

\noindent For the PC-based trainings of VGG9, the hyperparameter design space exploration has been focused on higher momentums for the SGD optimizer, instead of focusing on high inference rates. This decision has been taken to explore hyperparameters combinations that would favor the training stability, as VGG9 has been proven to be harder to train with PC.\\
In order to avoid abrupt changes in the test accuracy for the PC-based trainings, the training hyperparameters choice has been constrained to the ones for which, during the design space exploration, the following properties held:
\begin{itemize}
    \item After the 5th epoch, there was never a test accuracy decrease of no more than 5 percentage points between one epoch and the next.
    \item The final epoch accuracy was no lower than 5 points with respect to the maximum one.
\end{itemize}
Regarding the BP-based domain adaptation, a weights learning rate ($\theta_{lr}$) of $1e^{-3}$ has been used only for this training technique, since lower $\theta_{lr}$ values did not always provide a satisfactory test accuracy.\\
\noindent Finally, Table \ref{tab:datasets_and_transformations} summarizes the epochs, batch size, normalization values and data augmentation that were used for training from scratch and domain adapting, both using BP and PC.

\begin{table}
    \centering
    \caption{Summary of training epochs, batches, normalization values and data augmentations for the datasets used in the experiments.}
    \label{tab:datasets_and_transformations}
    \setlength{\tabcolsep}{2\tabcolsep} %
    \begin{tabularx}{\textwidth}{p{3cm} p{2cm} >{\raggedright\arraybackslash}X}
        \toprule
         & \textbf{MNIST} & \textbf{CIFAR10}\\
         \midrule
         Training Epochs & 10 & 100\\\midrule
         Batch size & 128 & 128\\\midrule
        \makecell[l]{Normalization\\mean} & - & [0.4914, 0.4822, 0.4465]\\\midrule
        \makecell[l]{Normalization\\standard deviation} & - & [0.2023, 0.1994, 0.2010]\\\midrule
        \makecell[l]{Data\\augmentation} & - & \makecell[l]{(4, 4) zero pad + random cropping 32x32,\\random (50\%) horizontal flip}\\
        \bottomrule
    \end{tabularx}
\end{table}

\end{document}